\documentclass[9pt, conference]{IEEEtran}
\IEEEoverridecommandlockouts
\usepackage{cite}
\usepackage{array}
\usepackage{amsmath,amssymb,amsfonts}
\usepackage{algorithmic}
\usepackage{graphicx}
\usepackage{textcomp}
\usepackage{xcolor}
\usepackage{multirow}
\usepackage{url}
\def\BibTeX{{\rm B\kern-.05em{\sc i\kern-.025em b}\kern-.08em
    T\kern-.1667em\lower.7ex\hbox{E}\kern-.125emX}}
\begin{document}
\bstctlcite{IEEEtran:BSTcontrol}

\title{Domain-Independent Automatic Generation of \\Descriptive Texts for Time-Series Data
}
% \name{Author Name$^{\star \dagger}$ \qquad Author Name$^{\star}$ \qquad Author Name$^{\dagger}$}

% \address{$^{\star}$ Affiliation Number One \\
% $^{\dagger}$}Affiliation Number Two

\author{\IEEEauthorblockN{
Kota Dohi$^{1}$, Aoi Ito$^{1, 2, *}$, Harsh Purohit$^{1}$, Tomoya Nishida$^{1}$, Takashi Endo$^{1}$, and Yohei Kawaguchi$^{1}$}
\IEEEauthorblockA{$^{1}$\textit{R\&D Group, Hitachi Ltd.} $^{2}$\textit{Hosei University}
\\
% Tokyo, Japan\\
\small\texttt{[kota.dohi.gr,
yohei.kawaguchi.xk]@hitachi.com} }
\thanks{\textit{*Aoi Ito has contributed to this work during an internship at Hitachi Ltd.}}
}

% }
\maketitle

% \author{\IEEEauthorblockN{Kota Dohi}
% \IEEEauthorblockA{%\textit{dept. name of organization (of Aff.)} \\
% \textit{Hitachi Ltd.}\\
% Tokyo, Japan \\
% kota.dohi.gr@hitachi.com}
% \and
% \IEEEauthorblockN{Aoi Ito}
% \IEEEauthorblockA{%\textit{dept. name of organization (of Aff.)} \\
% \textit{Hosei University}\\
% Tokyo, Japan \\
% aoi.ito.8q@stu.hosei.ac.jp}
% \and
% \IEEEauthorblockN{Yohei Kawaguchi}
% \IEEEauthorblockA{%\textit{dept. name of organization (of Aff.)} \\
% \textit{Hitachi Ltd.}\\
% Tokyo, Japan \\
% yohei.kawaguchi.xk@hitachi.com}
% }
% \vspace{1em}  
% \noindent\rule{\textwidth}{0.4pt} % 横線  

% \maketitle

\begin{abstract}
Due to scarcity of time-series data annotated with descriptive texts, training a model to generate descriptive texts for time-series data is challenging. In this study, we propose a method to systematically generate domain-independent descriptive texts from time-series data. We identify two distinct approaches for creating pairs of time-series data and descriptive texts: the forward approach and the backward approach. By implementing the novel backward approach, we create the Temporal Automated Captions for Observations (TACO) dataset. Experimental results demonstrate that a contrastive learning based model trained using the TACO dataset is capable of generating descriptive texts for time-series data in novel domains.
% To tackle this problem, we develop a In this study, we introduce explores a method to automatically generate descriptive texts for a wide range of time-series data. Previous research has mainly targeted specific domains, limiting their general applicability. To address this, we propose a domain-independent approach for generating detailed captions. Given the scarcity of domain-independent captioned time-series data, we introduce two datasets: the SUSHI dataset with automatically generated time-series data and captions, and the TACO dataset with automatically generated captions for real observed data. By using these datasets, our model learns to capture various characteristics of time-series data, and our proposed method has shown promising results in generating detailed captions for general time-series data.
% We developed a method for generating domain-independent descriptive texts from time-series data, addressing data scarcity by creating the TACO dataset with 1.2 million real samples. Using a contrastive learning approach, our model successfully generates descriptive texts for both in-domain and out-of-domain datasets. 
\end{abstract}


\section{Introduction}
\label{sec:intro}
The states of any system over time can be captured as time-series data. While time-series data provides comprehensive observational information about the system, generating descriptive texts from this data can significantly enhance human understanding \cite{kobayashi2013probabilistic, murakami2017learning}. 

Previous research on text generation from time-series data typically involves the development of domain-specific models by leveraging domain-specific knowledge \cite{andreas2014grounding, sowdaboina2014learning, banaee2013towards}. However, such specialized models lack the flexibility to generate text in novel domains. Conversely, with the recent advancements in large language models (LLMs), it has become feasible to generate plausible texts for unfamiliar domains by leveraging the commonsense knowledge embedded in pre-trained LLMs \cite{Zamfirescu2023, arora2023ask}. Nevertheless, solely relying on pre-trained LLMs often results in low accuracy when describing the characteristics of time-series data, necessitating annotated time-series data for training. The core issue is the scarcity of openly available, domain-independent annotated time-series data.

In this paper, we propose a method to systematically generate domain-independent descriptive texts from time-series data, addressing the issue of insufficient training data. Our contributions are threefold. First, based on our expertise and prior research, we establish a set of time-series classes that encapsulate the general characteristics of time-series data. Second, we describe two approaches for creating pairs of time-series data and classes, and for generating descriptive texts from these classes: the forward approach and the backward approach. The forward approach synthesizes pairs of time-series data and texts by inputting time-series classes and using corresponding functions (e.g., SUSHI dataset \cite{kawagu_sushi}). On the other hand, the backward approach assigns texts based on scores computed by predefined procedures for each time-series class, using real-world time-series data as input. Third, to validate the novel backward approach, we apply it to 1.2 million samples of real-world time-series data, creating the Temporal Automated Captions for Observations (TACO) dataset. Using the TACO dataset, we train a contrastive learning-based text generation model. Quantitative evaluation with caption evaluation metrics demonstrates that the trained model can generate domain-independent descriptive texts for time-series data in novel domains.

\section{Relation to prior work}

The task of generating descriptive texts from time-series data has been studied across various domains, including finance \cite{andreas2014grounding, murakami2017learning, spreafico2020neural}, meteorology \cite{sowdaboina2014learning}, and healthcare \cite{banaee2013towards}. These studies focus on a specific domain and generate descriptive texts by leveraging domain-specific knowledge. While such domain-specific models are effective within their respective domains, they fail to generate descriptive texts for data from other domains. In contrast, this study proposes a method for generating domain-independent descriptive texts from general time-series data, without relying on domain-specific knowledge.
% This approach allows for the generation of explanatory texts for general time-series data even in the absence of domain knowledge. Additionally, it can be utilized as a building block for producing final outputs when partial domain knowledge is available.

Several studies have proposed methods for explaining time-series data from image plots of the time-series data \cite{mahinpei2022linecap, spreafico2020neural}. However, in these studies, the generated explanations are either domain-specific or limited to a few basic characteristics of the time-series data. Additionally, when time-series data exhibit complex trends, converting them into images might result in the loss of important information. In contrast, this study proposes a method for explaining various characteristics of time-series data from one-dimensional sequential values of the time-series data.

\section{Approaches for generating pairs of time series data and corresponding descriptive texts}
As discussed in Section \ref{sec:intro}, there is a notable scarcity of datasets that pair time-series data with descriptive texts. Developing a model that can generate descriptive texts for diverse time-series data necessitates a substantial amount of time-series data annotated with descriptive texts. Therefore, we propose the automatic generation of pairs of time-series data and descriptive texts. 

To systematically generate descriptive texts for time-series data, we first define a set of time-series classes that represent various trends or characteristics inherent to time-series data. For example, a time-series class ``Rising'' can be assigned to any data exhibiting an upward trend, and a time-series class ``Dropout'' can be assigned to data showing sudden drops in values. The complete set of time-series classes are listed in Table \ref{tab:table1}. These classes were defined based on our expertise as signal processing engineers and a prior study by Imani et al. \cite{Imani2019}, which identified 20 local features that can be computed from time-series data.

\begin{table}[t]  
  \caption{List of time-series classes used for creating each dataset}  
  \centering  
  \begin{tabular}{|>{\raggedright\arraybackslash}m{0.045\textwidth}|>{\raggedright\arraybackslash}m{0.390\textwidth}|}  
    \hline  
    \textbf{Dataset} & \textbf{Time-series classes used for creating the dataset} \\  
    \hline  
    SUSHI & Constant, Linear increase / decrease, Concave, Convex, Exponential growth / decay, Inverted exponential growth / decay, Sigmoid, Inverted sigmoid, Cubic function, Negative cubic function, Gaussian, Inverted Gaussian, Sinusoidal wave, Square wave, Sawtooth wave, Reverse sawtooth wave, Triangle wave, Smooth, Noisy, Positive / Negative spiky, Positive and negative spiky, Steppy\\
    \hline
    TACO & Rising, Falling, Constant, Convex, Concave, Linear, Nonlinear, Smooth, Noisy, Simple, Complex, Spiky, Dropout, Periodic, Aperiodic, Symmetry, Asymemtry, Step, NoStep, High amplitude, Low amplitude\\
    \hline 
  \end{tabular}  
  \label{tab:table1}  
\end{table}

In the next step, we designed methods for generating pairs of time-series data and their corresponding time-series classes. We identified two distinct approaches for creating these pairs: the forward approach and the backward approach.

The forward approach involves generating both time-series data and descriptive texts based on time-series classes. The time-series data can be synthetically generated using functions that correspond to these classes, while the descriptive texts are derived from the names of the time-series classes. The SUSHI dataset was created using this approach. The time-series classes used for generating the SUSHI dataset are listed in Table \ref{tab:table1}. For detailed procedures on creating the dataset, please refer to the project page \cite{kawagu_sushi}.

The backward approach involves identifying time-series classes within time-series data and subsequently generating descriptive texts based on these identified classes. To facilitate the identification process, a method for calculating scores should be developed for each time-series class. Based on the score calculated for each time-series class and a threshold for each class, the appropriate time-series class is assigned to the corresponding time-series data. In this paper, we employed the backward approach to create the TACO dataset. The time-series classes used for generating this dataset are listed in Table \ref{tab:table1}. Detailed procedures for generating the TACO dataset will be described in a later section.

Both the forward and backward approaches have their advantages and disadvantages. In the forward approach, any named function can be used to create time-series data, allowing for the generation of data with a wide variety of functions. However, this approach cannot generate captions for real-world time-series data. Additionally, because the generated time-series data are synthetic, they may not encompass the diversity of time-series data observed in the real world. 
In the backward approach, captions can be generated for real data by identifying time-series classes within the observed data. However, methods for calculating scores corresponding to each time-series class must be implemented, and for some classes, this can be more challenging. For example, identifying that a time-series data should be assigned the class ``Rising'' can be easily done by calculating the correlation between the timestamps and the signal of that data. Conversely, determining whether a time-series data should be assigned the class ``Sigmoid'' is not as straightforward.
Therefore, we can conclude that both the forward and backward approaches are useful for generating pairs of time-series data and corresponding descriptive texts. If the score calculation for a time-series class can be performed using a combination of traditional signal processing techniques, that class can be utilized in both the forward and backward approaches. Otherwise, the class can be utilized only in the forward approach.

\begin{table}[t]  
  \caption{Time-Series Features and Their Calculation Methods}  
  \centering  
  \renewcommand{\arraystretch}{1.45}
  \begin{tabular}{|>{\raggedright\arraybackslash}m{0.075\textwidth}|>{\raggedright\arraybackslash}m{0.36\textwidth}|}  
    \hline  
    \textbf{Time-series class} & \textbf{Procedures for calculating scores and identifying the corresponding time-series class} \\  
    \hline  
    Rising / Falling & Higher / lower correlation between the timestamps and signals in time-series data. \\
    \hline
    Constant & Smaller average Wasserstein distance (WSD) between the segments of the time-series data.\\
    \hline  
    Convex / Concave & Larger average difference between the error of fitting linear curve and quadratic curve. Convex or Concave is determined by the sign of the coefficient of the quadratic term when fitting a quadratic curve.\\
    \hline  
    Linear / Nonlinear & Smaller / larger mean squared error when fitting a linear function.\\
    \hline
    Smooth & Smaller mean squared error between the signal obtained by taking the moving average of the signal and the original signal.\\  
    \hline  
    Noisy & Larger mean squared error between the signal obtained by applying a median filter to the difference between the median-filtered signal and the original signal.\\
    \hline  
    Simple / Complex & Smaller / larger average WSD between segments of a signal that is processed by taking the first derivative of the original signal.\\
    \hline  
    Spiky / Dropout & 
    Larger maximum difference between the original signal and a threshold. After segmenting the signal, the threshold is set for each segment. The threshold is obtained by adding (for ``Spiky'') or subtracting (for ``Dropout'') a value obtained by multiplying the standard deviation of the median filtered signal and a constant value to the median value of the segment.\\  
    \hline
    Periodic / Aperiodic & Smaller / larger difference between the error of fitting linear curve and the error between a periodic curve and the original signal. To create the periodic curve, first calculate the autocorrelation coefficient for the signal and detect the negative points of the second difference of the auto correlation coefficient as peaks. Then, create the periodic signal by treating the segment up to the peak point as one complete cycle and then repeating this cycle.\\
    \hline  
    Symmetry / Asymmetry & Smaller / larger minimum error between the padded original signals and their flipped signals. The padded original signals are generated by padding the signal at the front and back with various padding widths. The flipped signals are generated by left-right flipping each padded original signal.\\  
    \hline  
    Step / NoStep & Larger / smaller maximum value when convolving the original signal with kernels of different lengths. Each kernel has a value of $-1$ for the first half and a value of $1$ for the second half.\\  
    \hline  
    High amplitude / Low amplitude & Larger / smaller maximum variance for each segment of the original signal. \\  
    \hline  
    
  \end{tabular}  
  \label{tab:table2}  
\end{table}  

\section{Procedures for generation of the TACO dataset}
\label{sec:taco}
In this section, we detail the implementation of the backward approach in generating the TACO dataset. The TACO dataset was created from 1.2 million series of sensor data. For each series of data, time-series classes were assigned, and descriptive texts were generated based on these classes.

To identify time-series classes in time-series data, a score calculation method and a corresponding threshold were defined for each class. To ensure the applicability of the same score calculation methods and thresholds across different time-series datasets, we normalized both the input time-series data and the calculated scores. Specifically, the input time-series data were normalized between 0 and 1 using min-max scaling. Additionally, the calculated scores were normalized based on the number of data points in the time-series data.

The score calculation methods were designed using an iterative process. In the first step, scores were calculated for each time-series data using initially defined score calculation methods. In the next step, we inspected visual plots of the time-series data and the corresponding scores to update the calculation methods. This update ensures that the scores indicate the extent to which the characteristics of the corresponding time-series class are contained in each time-series data. By iterating the first and second step, score calculation methods were designed. The designed procedures for each class are outlined in Table \ref{tab:table2}.

The thresholds for assigning time-series classes were manually determined by the authors. Although these thresholds were set based on the consensus among the authors, they can vary depending on the individuals and the distribution of the dataset. A more sophisticated procedure for deciding thresholds will be investigated in future work. For detailed descriptions on the designed score calculation methods and specific values of the thresholds, please refer to the project page \cite{dohi_taco}.

After identifying time-series classes for each time-series data, we generated descriptive texts based on the combination of these classes. 
We first defined pairs of time-series classes and corresponding base captions. For example, for the class ``Rising'', the base caption is ``The signal has a rising trend''. Likewise, for ``Smooth'', the base caption is ``The signal has a smooth shape''. Then, we prepared base descriptive texts for each time-series data by combining the base captions corresponding to the time-series classes assigned to that data. For instance, for a time-series data with classes ``Rising'' and ``Smooth'', the base descriptive texts are ``The signal has a rising trend. The signal has a smooth shape''. Finally, we rephrased the base descriptive texts using the Llama-3-8B-Instruct \footnote{\url{https://huggingface.co/meta-llama/Meta-Llama-3-8B-Instruct}} \cite{dubey2024llama3herdmodels} to generate descriptive texts. 

% We then use LLM to rephrase the base caption and generate the final caption for each time series data. We used the Llama3 8B parameter model for rephrasing.

\section{Experiment}
\subsection{Model for generating descriptive texts from time-series data}
% We used a contrastive learning method for training the model. 
% To train a model for generating descriptive texts from using pairs of time-series data and descriptive texts. 
Inspired by prior works on caption generation for images \cite{mokady2021clipcap} and speech \cite{xu2024secap}, we trained our model using contrastive learning with a two-step training process.  
In the first step, we performed contrastive learning using the Informer \cite{zhou2021informer} as the signal encoder and the T5-small model \cite{raffel2020exploring} as the text encoder. These encoders have been selected due to their lightweight nature and their extensive utilization across their respective domains.  For each pair of time-series data and descriptive texts, the former was fed into the signal encoder and the latter into the text encoder. The goal of this step was to train an encoder model capable of producing shared embeddings for both time-series data and descriptive texts. During the training process, the parameters of the text encoder were frozen, and only the parameters of the signal encoder and the projection heads were updated. The projection heads, consisting of linear layers, were responsible for aligning the dimensions between the signal encoder and the text encoder.
%The input to the Informer consisted of 2048-dimensional vectors.

In the next step, we first calculated the embeddings of the time-series data using the encoder trained in the previous step. These embeddings were then transformed by a transformer-based bridge network so that they could be converted into descriptive texts by a text decoder. For the text decoder, we used T5-small. During this training step, the parameters of the entire encoder trained in the previous step and the text decoder were frozen, and only the parameters of the bridge network were trained.
%The bridge network is responsible for converting the embeddings obtained from the encoder into embeddings that can be handled by the text decoder.  

% In the following, we refer to this model as Contrastive Learning between Texts and Time series (CLTT).
For training the signal encoder, we used a batch size of 640 and a learning rate of $0.0001$. The bridge network was trained using a batch size of 1,280 and a learning rate of $0.00001$. For both training, AdamW optimizer \cite{loshchilov2019decoupled} was used.

\begin{table*}[t]  
\begin{center}  
\caption{Evaluation metrics for the generated descriptive texts. For all evaluation metrics, higher scores indicate a higher correspondence between the generated texts and the ground truth.  ``Proposed'' denotes the model trained using the contrastive learning based approach}  
\setlength\tabcolsep{8pt}
\begin{tabular}{@{\hspace{10pt}}l@{\hspace{10pt}}l@{\hspace{10pt}}l@{\hspace{10pt}}|@{\hspace{10pt}}c@{\hspace{10pt}}c@{\hspace{10pt}}c@{\hspace{10pt}}c@{\hspace{10pt}}c@{\hspace{10pt}}c@{\hspace{10pt}}c@{\hspace{10pt}}c@{\hspace{10pt}}}    
\hline
Train & Test    & Method&  {BLEU\_3} & {BLEU\_4} & {METEOR}& {ROUGE\_L} & {CIDEr} & {SPICE} &{BERTScore} &{Sentence BERT} \\     
\hline      
TACO & TACO   & NearNBR   &  0.214      &  0.139   & 0.182   & 0.347       & 0.090     & 0.128   & 0.883  & 0.802  \\      
TACO & TACO   & Proposed &  \textbf{0.224}      &  \textbf{0.145}        & \textbf{0.190}   & \textbf{0.356}       & \textbf{0.094}     & \textbf{0.126}   & \textbf{0.892} & \textbf{0.827}  \\ \hline   

TACO & Weather   & NearNBR &  0.227      &  0.149        & 0.187   & 0.352      & 0.097       & 0.126   & 0.878 & 0.793   \\     
TACO & Weather & Proposed      &  \textbf{0.320}      &  \textbf{0.221}       & \textbf{0.226}   & \textbf{0.421}      & \textbf{0.240}       & \textbf{0.170}   & \textbf{0.882} & \textbf{0.796}    \\  \hline   

TACO & ETTh1   & NearNBR &  0.181      &  0.121        & 0.163   & 0.319      & 0.063       & 0.108   & 0.852 & 0.711   \\     
TACO & ETTh1 & Proposed      &  \textbf{0.269}      &  \textbf{0.189}       & \textbf{0.239}   & \textbf{0.413}      & \textbf{0.229}       & \textbf{0.167}   & \textbf{0.880} & \textbf{0.796}    \\  \hline  

TACO & Electricity & NearNBR &  0.205      &  0.135        & 0.173   & 0.329      & 0.070       & 0.119   & 0.855 & 0.745   \\     
TACO & Electricity & Proposed      &  \textbf{0.259}      &  \textbf{0.175}       & \textbf{0.238}   & \textbf{0.393}      & \textbf{0.153}       & \textbf{0.186}   & \textbf{0.878} & \textbf{0.807}    \\   \hline  

TACO & Exchange   & NearNBR &  0.246     &  0.172        & 0.216   & 0.390      & 0.186       & 0.151   & 0.872 & 0.776   \\     
TACO & Exchange & Proposed      &  \textbf{0.355}      &  \textbf{0.261}       & \textbf{0.249}   & \textbf{0.454}      & \textbf{0.344}       & \textbf{0.169}   & \textbf{0.890} & \textbf{0.813}    \\   \hline   

TACO & Traffic   & NearNBR &  0.157      &  0.099        & 0.163   & 0.314      & 0.055       & 0.103   & 0.849 & 0.723   \\     
TACO & Traffic & Proposed      &  \textbf{0.257}      &  \textbf{0.176}       & \textbf{0.233}   & \textbf{0.404}      & \textbf{0.172}       & \textbf{0.163}   & \textbf{0.879} & \textbf{0.827}    \\   

\hline  
\end{tabular}  
\label{tab:result_1}  
\end{center}  
\end{table*}  

\begin{table*}[]
\centering
\caption{Examples of descriptive texts generated from the model}
\renewcommand{\arraystretch}{1.5} % Adjusts the row height
\setlength{\tabcolsep}{6pt}       % Adjusts the space between columns
\begin{tabular}{m{3cm} m{0.80cm} m{3cm} m{4.5cm} m{4.5cm} }
\hline
\begin{tabular}{r}\textbf{Train}\end{tabular} & \textbf{Test} & \textbf{Time-series data} & \textbf{Ground truth using backward approach} & \textbf{Generated texts} \\
\hline
% Row 1
\begin{tabular}{l}TACO \\(backward)\end{tabular}&
TACO&
\vspace{0.2pt}\raisebox{-0.5\height}{\includegraphics[width=\linewidth]{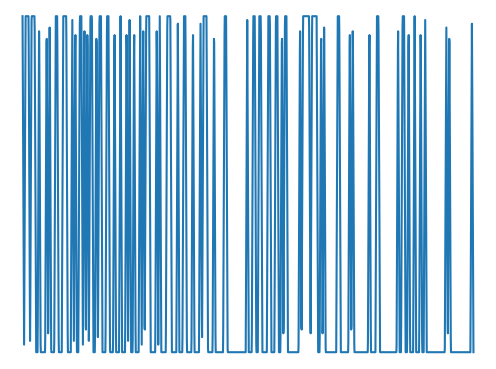}}\vspace{0.2pt} & 
The time series signal exhibits a non-linear trend, complex behavior, and is heavily noisy, featuring sudden spikes and irregular patterns, with no clear periodicity, and displaying asymmetrical characteristics, further characterized by a high amplitude.& 
The signal exhibits a non-linear progression, accompanied by significant noise and complexity, featuring sudden spikes and irregular patterns, with an asymmetrical structure and a high amplitude.\\
\hline

% Row 2
\begin{tabular}{l}TACO \\(backward)\end{tabular}&
Weather&
\vspace{0.2pt}\raisebox{-0.5\height}{\includegraphics[width=\linewidth]{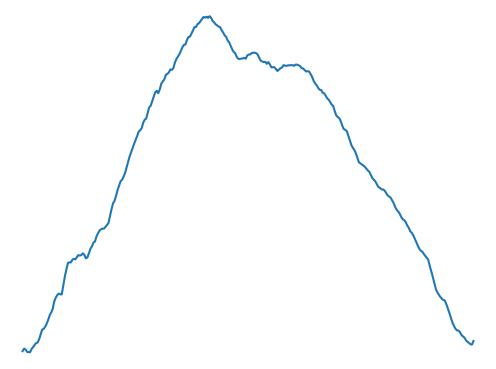}}\vspace{0.2pt} & 

A time series signal with a concave shape, exibiting a nonlinear trend, simplicity, and aperiodic behavior, showcasing symmetry throughout its oscillations.& 
The signal exhibits a concave shape, with a nonlinear trend that is smooth and simple, displaying aperiodic and symmetric patterns. \\
\hline

% Row 3
\begin{tabular}{l}SUSHI  + TACO\\ (forward \& backward)\end{tabular}&
% \raisebox{2\height}{SUSHI} \\
% \raisebox{3\height}{+ TACO} \\
% \raisebox{3\height}{(forward} \\
% \& backward)\end{tabular}&
Weather&
\vspace{0.2pt}\raisebox{-0.5\height}{\includegraphics[width=\linewidth]{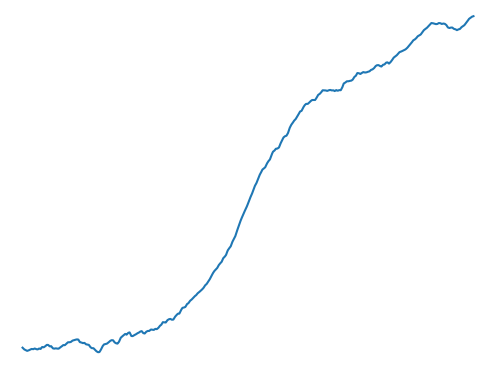}}\vspace{0.2pt} & 
% \raisebox{-0.5\height}
A time series signal exhibiting a rising trend, smooth shape, simplicity, aperiodicity, and asymmetry.& 
This trend shows a steadily increasing trajectory, similar to a sigmoid curve, approaching a value, with a decreasing rate of increase. The signal exhibits a steady upward progression, characterized by a linear trajectory, and maintains a smooth, uncomplicated shape, lacking sharp changes or abrupt shifts. \\
\hline
% Additional rows as needed
\end{tabular}
\label{tab:spectrograms}
\end{table*}

\subsection{Evaluation}
We conducted three experiments to validate that a model can be trained to generate domain-independent descriptive texts for general time-series data. The experiments utilized two sets of datasets. The first set is the in-domain datasets, which include the SUSHI dataset and TACO dataset. The SUSHI dataset consists of 1.4 million synthetic pairs of time-series data and descriptive texts generated using a forward approach, with each time-series having a length of 2,048 points. The TACO dataset was created using the backward approach described in Section \ref{sec:taco}. This dataset includes 1.2 million real-world sensor data series, for which time-series classes were identified, and descriptive texts were generated based on these classes. Each time-series in the TACO dataset originally had a length of 300 points. To ensure consistency, all time-series data were interpolated to 2,048 points using linear interpolation.

The second set is the out-of-domain (OOD) datasets, which consist of five publicly available time-series datasets (Weather, ETTh1, Electricity, Exchange, Traffic) provided by Wu et al. \cite{wu2021autoformer}. For each of these datasets, the time-series data from each column were segmented into 300-point sections, and each segment was interpolated to 2,048 points using linear interpolation. Since the OOD datasets lack descriptive texts, we employed the backward approach to generate ground truth texts for these datasets. For each OOD dataset, 100 samples are randomly sampled to be used for the evaluation.
%Ground-truth descriptive texts for these datasets were prepared using the same procedures as the TACO dataset.

In the first experiment, we investigated whether the trained model can generate descriptive texts for in-domain datasets. Given that generating texts for real sensor data is more challenging, we used the training subset of the TACO dataset for training and the test subset of the TACO dataset for evaluation.

In the second experiment, we explored whether the trained model can generate descriptive texts for out-of-domain (OOD) datasets.  We used the training subset of the TACO dataset for training the model and the OOD datasets for evaluation.

For both the first and second experiments, we conducted quantitative evaluations of the generated texts using automated caption metrics, including BLEU\_3 \cite{papineni2002bleu}, BLEU\_4, METEOR \cite{lavie2007meteor}, ROUGE\_L \cite{lin2004rouge}, CIDEr \cite{vedantam2015cider}, SPICE \cite{anderson2016spice}, BERTScore \cite{zhang2020bertscore}, and Sentence-BERT \cite{reimers2019sentence}.
% We conducted three experiments for evaluation. 
We used NearNBR \cite{jhamtani-berg-kirkpatrick-2021-truth} as the baseline method. This method calculates the mean squared error between the input time-series data and each time-series data in the training dataset, and then outputs the time-series data and corresponding texts with the minimum mean squared error.

% for each test time-series data, this method outputs time-series data and corresponding texts with minimum mean squaredcalculates the mean squared error calculated the mean squared errorThis method selects the time-series data in the training dataset with the smallest mean squared error is selected, and the corresponding descriptive texts are output.
In the third experiment, we examined whether a model can be trained to leverage descriptive texts generated by the forward approach and those generated by the backward approach. Specifically, we generated descriptive texts for time-series data in the SUSHI dataset using the backward approach. These texts were then concatenated with the original descriptive texts in the SUSHI dataset, which were created using the forward approach. The goal of this concatenation was to ensure that the output texts from the trained model would reflect the characteristics of both approaches.

\subsection{Results}
In Table \ref{tab:result_1}, the quantitative evaluation results for the generated texts are presented. The model trained on the TACO dataset consistently outperforms the NearNBR model. Both the BERTScore (a word-level metric) and the Sentence BERT score (a sentence-level metric) tend to degrade when evaluated on out-of-domain (OOD) datasets compared to the in-domain TACO dataset. This degradation can be attributed to the differences in characteristics between the training dataset (TACO) and the test OOD datasets. Specifically, the NearNBR model exhibited a maximum degradation of 0.034 in BERTScore and 0.091 in Sentence BERT score. In contrast, the contrastive learning-based model showed a maximum degradation of 0.014 in BERTScore and 0.031 in Sentence BERT score. These findings indicate that while the NearNBR model is significantly affected by the characteristic differences between the in-domain and OOD datasets, our trained model effectively mitigates the degradation caused by these differences. In conclusion, the trained model successfully captured time-series characteristics defined in the TACO dataset and generated descriptive texts for both in-domain and OOD datasets.

In Table \ref{tab:spectrograms}, examples of descriptive texts generated by the trained model are shown. The generated descriptive texts successfully represent various features inherent in the time-series data. 
% When trained the model using both SUSHI and TACO dataset, the generated descriptive texts of time-series data from OOD datasets reflected both the texts in the SUSHI and those in the TACO dataset. ``Rising''
In the third row of Table \ref{tab:spectrograms}, an example of texts generated by a model trained using both the SUSHI dataset and TACO dataset are shown. The first sentence of the generated text represents the descriptive texts generated using the forward approach (time-series class ``Sigmoid'' is only contained in the SUSHI dataset), while the second sentence represents the descriptive texts generated using the backward approach. Compared to the ground truth texts generated using only the backward approach, the sentences generated by the trained model capture more diverse characteristics of the time-series data. This observation indicates that the joint use of the forward and backward approaches contributes to generating more diverse descriptive texts.

\section{Conclusion}
To handle scarcity of time-series data annotated with descriptive texts, we proposed a method for systematically generating descriptive texts from time-series data. We first defined time-series classes and identified the forward approach and backward approach. We then implemented the novel backward approach to generate descriptive texts for 1.2 million sensor time-series, and created the TACO dataset. 
We experimentally verified that a contrastive learning model using the TACO dataset can generate descriptive texts for time-series data in novel domains.
We also demonstrated that the forward and backward approach can be jointly used for generating datasets for training the model.
% By training a contrastive learning based model using the TACO datsaet and generating descriptive texts for time-series data, we experimentally verified that 
% We also used the SUSHI dataset
% We explored a method for generating descriptions for general time series data without using domain knowledge. While traditional description generation models in previous research assumed the existence of domain knowledge, our proposed method enables the explanation of general time series shapes and trends even in the absence of domain knowledge. To create the training data for the model that generates descriptions for time series data, we developed a method to automatically generate captions for real data. By performing contrastive learning using the created caption and time series pairs, it was demonstrated that a model can be trained to explain various features of general time series.

% We developed a method for generating domain-independent descriptive texts from time-series data, addressing data scarcity by creating the TACO dataset with 1.2 million real samples. Using a contrastive learning approach, our model successfully generates descriptive texts for both in-domain and out-of-domain datasets. 

\bibliographystyle{IEEEtran}
\bibliography{refs}

% Generated by IEEEtran.bst, version: 1.12 (2007/01/11)
\begin{thebibliography}{10}
\providecommand{\url}[1]{#1}
\csname url@samestyle\endcsname
\providecommand{\newblock}{\relax}
\providecommand{\bibinfo}[2]{#2}
\providecommand{\BIBentrySTDinterwordspacing}{\spaceskip=0pt\relax}
\providecommand{\BIBentryALTinterwordstretchfactor}{4}
\providecommand{\BIBentryALTinterwordspacing}{\spaceskip=\fontdimen2\font plus
\BIBentryALTinterwordstretchfactor\fontdimen3\font minus \fontdimen4\font\relax}
\providecommand{\BIBforeignlanguage}[2]{{%
\expandafter\ifx\csname l@#1\endcsname\relax
\typeout{** WARNING: IEEEtran.bst: No hyphenation pattern has been}%
\typeout{** loaded for the language `#1'. Using the pattern for}%
\typeout{** the default language instead.}%
\else
\language=\csname l@#1\endcsname
\fi
#2}}
\providecommand{\BIBdecl}{\relax}
\BIBdecl

\bibitem{kobayashi2013probabilistic}
M.~Kobayashi, I.~Kobayashi, H.~Asoh, and S.~Guadarrama, ``A probabilistic approach to text generation of human motions extracted from kinect videos,'' in \emph{Proceedings of the World Congress on Engineering and Computer Science}, 2013.

\bibitem{murakami2017learning}
S.~Murakami \emph{et~al.}, ``Learning to generate market comments from stock prices,'' in \emph{Proceedings of the 55th Annual Meeting of the Association for Computational Linguistics (ACL 2017)}, 2017.

\bibitem{andreas2014grounding}
J.~Andreas and D.~Klein, ``Grounding language with points and paths in continuous spaces,'' in \emph{Proceedings of the Eighteenth Conference on Computational Natural Language Learning (CoNLL 2014)}, 2014.

\bibitem{sowdaboina2014learning}
P.~K.~V. Sowdaboina, S.~Chakraborti, and S.~Sripada, ``Learning to summarize time series data,'' in \emph{Proceedings of the International Conference on Intelligent Text Processing and Computational Linguistics (CICLING 2014)}, 2014.

\bibitem{banaee2013towards}
H.~Banaee, M.~U. Ahmed, and A.~Loutfi, ``Towards {NLG} for physiological data monitoring with body area networks,'' in \emph{Proceedings of the 14th European Workshop on Natural Language Generation}.\hskip 1em plus 0.5em minus 0.4em\relax Association for Computational Linguistics, 2013, pp. 193--197.

\bibitem{Zamfirescu2023}
J.~Zamfirescu-Pereira, R.~Y. Wong, B.~Hartmann, and Q.~Yang, ``Why johnny can’t prompt: How non-{AI} experts try (and fail) to design llm prompts,'' in \emph{Proceedings of the 2023 CHI Conference on Human Factors in Computing Systems}, ser. CHI '23.\hskip 1em plus 0.5em minus 0.4em\relax New York, NY, USA: Association for Computing Machinery, 2023.

\bibitem{arora2023ask}
S.~Arora and A.~Naray, ``Ask me anything: A simple strategy for prompting language models,'' in \emph{Proceedings of the International Conference on Learning Representations (ICLR)}, 2023.

\bibitem{kawagu_sushi}
Y.~Kawaguchi, K.~Dohi, and A.~Ito, ``{SUSHI}: {A} dataset of synthetic unichannel signals based on heuristic implementation,'' \url{https://github.com/y-kawagu/SUSHI}, 2024, accessed: 2025-08-05.

\bibitem{spreafico2020neural}
A.~Spreafico and G.~Carenini, ``Neural data-driven captioning of time-series line charts,'' in \emph{Proceedings of the 2020 International Conference on Advanced Visual Interfaces (AVI '20)}, 2020, pp. 1--5.

\bibitem{mahinpei2022linecap}
A.~Mahinpei, Z.~Kostic, and C.~Tanner, ``{LineCap}: Line charts for data visualization captioning models,'' in \emph{2022 IEEE Visualization and Visual Analytics (VIS)}, 2022, pp. 35--39.

\bibitem{Imani2019}
S.~Imani, S.~Alaee, and E.~Keogh, ``Putting the human in the time series analytics loop,'' in \emph{Companion Proceedings of The 2019 World Wide Web Conference}.\hskip 1em plus 0.5em minus 0.4em\relax Association for Computing Machinery, 2019, p. 635–644.

\bibitem{dohi_taco}
K.~Dohi, A.~Ito, and Y.~Kawaguchi, ``{TACO}: {Temporal} automated captions for observations,'' \url{https://github.com/Kota-Dohi/TACO}, 2024, accessed: 2025-08-05.

\bibitem{dubey2024llama3herdmodels}
A.~Dubey \emph{et~al.}, ``The {Llama} 3 herd of models,'' \emph{arXiv preprint arXiv:2407.21783}, 2024.

\bibitem{mokady2021clipcap}
R.~Mokady, A.~Hertz, and A.~H. Bermano, ``{ClipCap}: {CLIP} prefix for image captioning,'' \emph{arXiv preprint arXiv:2111.09734}, 2021.

\bibitem{xu2024secap}
Y.~Xu \emph{et~al.}, ``{SEcap}: Speech emotion captioning with large language model,'' in \emph{Proceedings of the AAAI Conference on Artificial Intelligence}, vol.~38, no.~17, 2024, pp. 19\,323--19\,331.

\bibitem{zhou2021informer}
H.~Zhou \emph{et~al.}, ``Informer: Beyond efficient transformer for long sequence time-series forecasting,'' in \emph{Proceedings of the AAAI Conference on Artificial Intelligence}, vol.~35, no.~12, 2021, pp. 11\,106--11\,115.

\bibitem{raffel2020exploring}
C.~Raffel \emph{et~al.}, ``Exploring the limits of transfer learning with a unified text-to-text transformer,'' \emph{Journal of Machine Learning Research}, vol.~21, no. 140, pp. 1--67, 2020.

\bibitem{loshchilov2019decoupled}
I.~Loshchilov and F.~Hutter, ``Decoupled weight decay regularization,'' in \emph{International Conference on Learning Representations (ICLR)}, 2019.

\bibitem{wu2021autoformer}
H.~Wu, J.~Xu, J.~Wang, and M.~Long, ``Autoformer: Decomposition transformers with auto-correlation for long-term series forecasting,'' in \emph{Advances in Neural Information Processing Systems (NeurIPS)}, 2021.

\bibitem{papineni2002bleu}
K.~Papineni, S.~Roukos, T.~Ward, and W.-J. Zhu, ``{BLEU}: a method for automatic evaluation of machine translation,'' in \emph{Proceedings of the 40th Annual Meeting of the Association for Computational Linguistics}.\hskip 1em plus 0.5em minus 0.4em\relax Association for Computational Linguistics, 2002, pp. 311--318.

\bibitem{lavie2007meteor}
A.~Lavie and A.~Agarwal, ``{METEOR}: {An} automatic metric for mt evaluation with high levels of correlation with human judgments,'' in \emph{Proceedings of the Second Workshop on Statistical Machine Translation}.\hskip 1em plus 0.5em minus 0.4em\relax Association for Computational Linguistics, 2007, pp. 228--231.

\bibitem{lin2004rouge}
C.-Y. Lin, ``{ROUGE}: {A} package for automatic evaluation of summaries,'' in \emph{Text Summarization Branches Out: Proceedings of the ACL-04 Workshop}.\hskip 1em plus 0.5em minus 0.4em\relax Association for Computational Linguistics, 2004, pp. 74--81.

\bibitem{vedantam2015cider}
R.~Vedantam, C.~L. Zitnick, and D.~Parikh, ``{CIDEr}: {Consensus-based} image description evaluation,'' in \emph{2015 IEEE Conference on Computer Vision and Pattern Recognition (CVPR)}.\hskip 1em plus 0.5em minus 0.4em\relax IEEE, 2015, pp. 4566--4575.

\bibitem{anderson2016spice}
P.~Anderson, B.~Fernando, M.~Johnson, and S.~Gould, ``{SPICE}: {Semantic} propositional image caption evaluation,'' in \emph{Proceedings of the European Conference on Computer Vision (ECCV)}.\hskip 1em plus 0.5em minus 0.4em\relax Springer, 2016, pp. 382--398.

\bibitem{zhang2020bertscore}
T.~Zhang, V.~Kishore, F.~Wu, K.~Q. Weinberger, and Y.~Artzi, ``{BERTScore}: {Evaluating} text generation with bert,'' in \emph{International Conference on Learning Representations (ICLR)}, 2020.

\bibitem{reimers2019sentence}
N.~Reimers and I.~Gurevych, ``{Sentence-BERT}: {Sentence} embeddings using siamese bert-networks,'' in \emph{Proceedings of the 2019 Conference on Empirical Methods in Natural Language Processing and the 9th International Joint Conference on Natural Language Processing (EMNLP-IJCNLP)}.\hskip 1em plus 0.5em minus 0.4em\relax Association for Computational Linguistics, 2019, pp. 3982--3992.

\bibitem{jhamtani-berg-kirkpatrick-2021-truth}
H.~Jhamtani and T.~Berg-Kirkpatrick, ``Truth-conditional captions for time series data,'' in \emph{Proceedings of the 2021 Conference on Empirical Methods in Natural Language Processing}, M.-F. Moens, X.~Huang, L.~Specia, and S.~W.-t. Yih, Eds.\hskip 1em plus 0.5em minus 0.4em\relax Online and Punta Cana, Dominican Republic: Association for Computational Linguistics, Nov. 2021, pp. 719--733.

\end{thebibliography}
\end{document}